\documentclass{article}

\usepackage[main, final]{neurips_2026}
\usepackage[utf8]{inputenc} % allow utf-8 input
\usepackage[T1]{fontenc}    % use 8-bit T1 fonts
\usepackage{hyperref}       % hyperlinks
\usepackage{url}            % simple URL typesetting
\usepackage{booktabs}       % professional-quality tables
\usepackage{amsfonts}       % blackboard math symbols
\usepackage{amsmath}        % math environments
\usepackage{amssymb}        % math symbols
\usepackage{mathtools}      % extended math tools
\usepackage{nicefrac}       % compact symbols for 1/2, etc.
\usepackage{microtype}      % microtypography
\usepackage{xcolor}         % colors
\usepackage{graphicx}       % images
\usepackage{subcaption}     % subfigures
\usepackage{algorithm}      % algorithm environment
\usepackage{algorithmic}    % algorithmic pseudocode
\usepackage{multirow}       % multi-row tables
\usepackage{colortbl}       % colored table rows
\usepackage{xspace}         % smart spacing after macros
\usepackage{enumitem}       % customizable lists
\usepackage{wrapfig}        % wrapped figures
\usepackage{tikz}
\usepackage{utfsym}
\usepackage{bm}

\newcommand{\pare}{\textsc{PARE}\xspace}
\definecolor{tablehl}{RGB}{230, 240, 255}

\title{PARE: Pruning and Adaptive Routing for Efficient Video Generation}

\author{%
  Yutong Wang\textsuperscript{1},
  Yunke Wang\textsuperscript{1},
  Tianfan Xue\textsuperscript{3,2},
  Yu Qiao\textsuperscript{2},
  Yaohui Wang\textsuperscript{2},\\
  \textbf{Xinyuan Chen\textsuperscript{2}\footnotemark[1]} \ ,
  \textbf{Chang Xu\textsuperscript{1}\thanks{Corresponding author.}}\\
  \textsuperscript{1}School of Computer Science, The University of Sydney, Australia\ \\
    \textsuperscript{2}Shanghai Artificial Intelligence Laboratory, China \ \\
    \textsuperscript{3}Multimedia Laboratory, The Chinese University of Hong Kong, China
}

\begin{document}

\maketitle
\begin{abstract}
Video Diffusion Transformers (DiTs) generate high-quality videos but demand substantial compute due to wide blocks, deep architectures, and iterative sampling.
Recent methods reduce cost by compressing width, depth, or sampling steps, but typically commit to a fixed architecture that cannot adapt to individual inputs or denoising stages.
We propose \textbf{PARE} (\textbf{P}runing and \textbf{A}daptive \textbf{R}outing for \textbf{E}fficient video generation), which jointly compresses width and depth with structure-aware pruning and input-adaptive routing.
For width, we observe that attention heads specialize into spatial and temporal roles, and design importance scoring that accounts for this distinction to prevent motion-critical temporal heads from being pruned prematurely.
For depth, we train a lightweight router conditioned on denoising timestep and visual content to dynamically select which blocks to execute at each step, enabling per-input compute adaptation rather than static block removal.
A progressive pipeline first recovers width-pruned quality via distillation, then jointly optimizes the student and router to decouple the two learning objectives.
Experiments on Wan2.1-14B for both image-to-video and text-to-video generation show that \pare substantially reduces per-step computation while preserving quality across VBench dimensions, and composes with step distillation for further acceleration.
Code and models are available at \url{https://github.com/hhhh1138/PARE}.
\end{abstract}

\section{Introduction}
\label{sec:intro}

Diffusion Transformers~\cite{dit} have brought video generation to remarkable fidelity~\cite{wan2025wan, yang2024cogvideox, hunyuanvideo, he2026posegen, li2026toward, yu2025core3d}, yet their inference cost remains prohibitive. Each denoising step must traverse dozens of transformer blocks with billions of parameters, and generating a single video requires tens of such steps. This cost spans three axes: \emph{width}, the heads and FFN dimensions within each block; \emph{depth}, the number of blocks executed per step; and \emph{steps}, the denoising iterations. Recent methods have begun to compress multiple axes jointly~\cite{diffpruning,karnewar2025neodragon,fastlightgen, wu2025mobilevideodit}, yet they remain \emph{static}: the same reduced architecture serves every input at every timestep, all attention heads are treated uniformly regardless of their functional role, and there is no mechanism to adapt computation to input complexity.

Two properties of video DiTs suggest a better strategy. First, attention heads exhibit clear spatial-temporal specialization: shallow heads attend predominantly within frames while deeper heads capture cross-frame motion. Magnitude-based pruning systematically undervalues temporal heads, whose activations are smaller yet critical for motion coherence. Second, block importance varies across both timesteps and inputs: early denoising steps that recover global structure rely on different blocks than late steps that refine details, and the optimal block subset further depends on scene complexity. These observations call for \emph{spatial-temporal aware width pruning} within blocks and \emph{content-adaptive depth routing} across blocks.

We present \textbf{PARE} (\textbf{P}runing and \textbf{A}daptive \textbf{R}outing for \textbf{E}fficient video generation), a framework that compresses video DiTs along both axes. For width, \pare scores attention heads via calibration-based importance with a protection multiplier for temporal heads, and selects FFN neurons through importance-plus-diversity filtering that exploits the factored up/down projection structure. For depth, a lightweight router conditioned on both timestep and visual content selects which blocks to execute, with a budget schedule that allocates more blocks to high-noise steps. A progressive training pipeline first recovers width-pruned quality via distillation, then jointly optimizes the pruned model and router, avoiding the compound difficulty of learning both from scratch.

We evaluate \pare on Wan2.1-14B for both image-to-video and text-to-video generation. \pare halves per-step computation while preserving quality across VBench~\cite{vbench} dimensions, and composes with step distillation for further acceleration along the third axis. Our contributions are:
\begin{itemize}[leftmargin=*, itemsep=2pt, topsep=2pt]
    \item A spatial-temporal aware width pruning strategy combining temporal head protection with importance-plus-diversity FFN selection that exploits the factored projection structure.
    \item A content-adaptive block router that conditions on both timestep and visual input to select which blocks to execute, with a budget schedule that allocates more computation to high-noise steps, enabling per-sample depth adaptation at negligible parameter overhead.
    \item A progressive training pipeline that decouples width distillation from joint width-routing optimization, breaking the circular dependency between pruned feature quality and routing decisions.
    \item Experiments on Wan2.1-14B showing that \pare halves per-step computation with minimal quality loss on both I2V and T2V, and up to ${\sim}50\times$ total speedup with step distillation.
\end{itemize}

\section{Related Work}
\label{sec:related}

\paragraph{Efficient Video Generation.}
Efficient video generation methods reduce inference cost through numerical precision~\cite{viditq,svdquant,wang2026pulsequant}, attention sparsity~\cite{svg,svg2,liu2026losa,li2026sol}, model compression~\cite{fastlightgen,karnewar2025neodragon}, and sampling acceleration~\cite{dmd, dmd2, vdot, wang2026vdotpp, chen2026stdec,wu2026d}.
Quantization lowers the precision of weights and activations: ViDiT-Q~\cite{viditq} studies quantization for image and video diffusion transformers, while SVDQuant~\cite{svdquant} uses a high-precision low-rank branch to absorb outliers and enable 4-bit diffusion inference.
Sparse attention reduces token interactions within each denoising step; Sparse VideoGen~\cite{svg} exploits spatial-temporal sparsity in video attention, and Sparse VideoGen2~\cite{svg2} improves sparse computation through semantic-aware permutation.
Structural pruning reduces network size by removing redundant components, while consistency models~\cite{consistency, lcm} and distribution matching distillation~\cite{dmd, dmd2, vdot, wang2026vdotpp, ge2026salt, ge2026saltpp} reduce the number of sampling steps.
These approaches address different sources of inference cost.
\pare focuses on reducing per-step computation through width pruning and adaptive depth routing, and combines this compression with step distillation for further acceleration.

\paragraph{Structural Pruning for Diffusion Models.}
Structural pruning~\cite{diffpruning,karnewar2025neodragon,wu2025mobilevideodit,hong2026safe,yu2026silsa,fastlightgen} compresses pretrained networks by removing channels, attention heads, FFN dimensions, or entire blocks.
For diffusion models, Diff-Pruning~\cite{diffpruning} aggregates Taylor-based importance estimates over selected timesteps to identify removable parameters.
Recent video generation methods combine structural compression with distillation to recover quality.
NeoDragon~\cite{karnewar2025neodragon} prunes MMDiT blocks according to their relative importance and restores performance through two-stage distillation.
MobileDiT~\cite{wu2025mobilevideodit} introduces sensitivity-aware tri-level pruning together with adversarial step distillation, while FastLightGen~\cite{fastlightgen} jointly addresses model-size and sampling-step reduction.
These studies demonstrate the effectiveness of combining structural pruning with distillation.
\pare further explores which intra-block components to retain, combining spatial-temporal aware head pruning with FFN selection.

\paragraph{Dynamic Computation for Diffusion Models.}
Dynamic computation~\cite{dydit,deepcache,yu2026elsa3d,liu2026palm,pab} allocates resources according to the requirements of individual inputs or denoising stages.
DyDiT~\cite{dydit} explores this direction through timestep-wise dynamic width and spatial-wise dynamic token selection, adapting computation within image DiTs.
A complementary line of work exploits redundancy across denoising steps through feature reuse.
DeepCache~\cite{deepcache} reuses high-level features across adjacent steps, while PAB~\cite{pab} reuses attention outputs over subsequent steps in video generation.
Routing and caching therefore address different aspects of redundant computation: routing selects the components to execute, whereas caching reuses previously computed representations.
\pare performs dynamic selection at block granularity within a width-pruned video DiT.
A timestep-dependent budget determines how many blocks are active, and a router conditioned on both timestep and visual content determines which blocks to execute.
This design enables per-sample depth adaptation within a single compressed model.

% \paragraph{Static Compression for Diffusion Models.}
% Recent video DiTs~\cite{shen2026fine,shen2026decoding,li2026unified} scale to billions of parameters, making per-step inference costly. Static compression reduces this cost through two orthogonal axes. \emph{Structural pruning} removes entire heads, neurons, or layers: WANDA~\cite{wanda} combines activations with weight norms for one-shot LLM pruning; Diff-Pruning~\cite{diffpruning} adapts Taylor scores to U-Net denoisers; for video DiTs, FastLightGen~\cite{fastlightgen} drops blocks via reconstruction error, NeoDragon~\cite{karnewar2025neodragon} removes MMDiT blocks with two-stage distillation, and Wu~\etal~\cite{wu2025mobilevideodit} search tri-level binary masks over 20K iterations. \emph{Step distillation} compresses the sampling trajectory: progressive distillation~\cite{progressivedistill} halves steps iteratively, consistency models~\cite{consistency, lcm} learn direct mappings, and distribution matching~\cite{dmd, dmd2, vdot, wang2026vdotpp, ge2026saltpp} closes the few-step gap. Both axes apply the same reduced model to every input and timestep. \pare complements them with spatial-temporal aware width pruning that protects motion-critical heads, dynamic depth routing that adapts per sample, and compatibility with step distillation for joint acceleration along all three axes.

\section{Method}
\label{sec:method}

\begin{figure}[t]
    \centering
    \includegraphics[width=\linewidth]{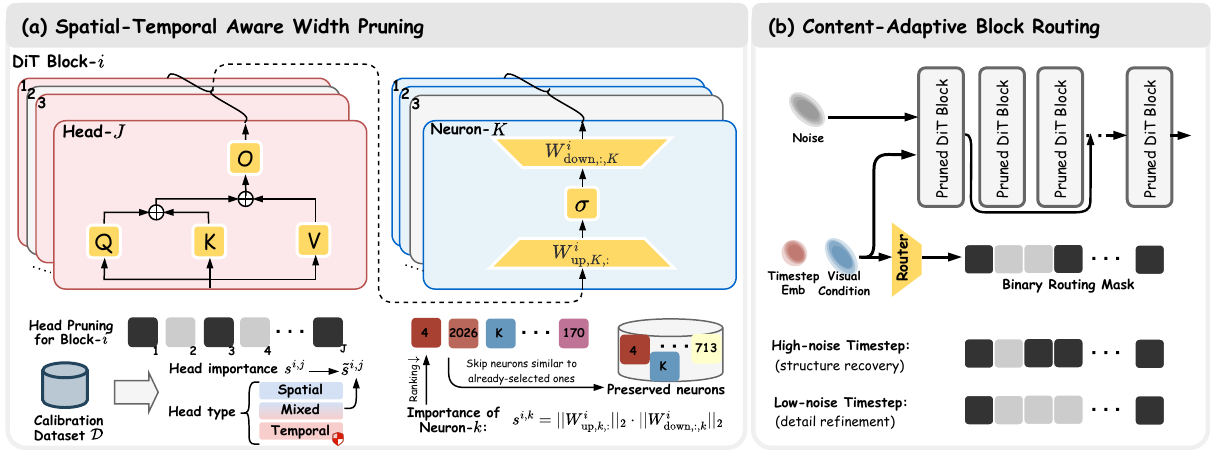}
    \caption{\textbf{Overview of \pare.}
    \textbf{(a)}~Spatial-temporal aware width pruning. 
    Calibration-based head importance scores $s^{i,j}$ are adjusted to $\tilde{s}^{i,j}$ according to head type, protecting temporal heads from magnitude bias. 
    FFN neurons are ranked by the product of their up/down projection norms and greedily selected to preserve importance while avoiding functional redundancy.
    \textbf{(b)}~Content-adaptive block routing. 
    A lightweight router conditioned on the timestep embedding and visual input produces a binary mask selecting which pruned blocks to execute. 
    A timestep-dependent budget assigns more blocks to high-noise steps for structure recovery and fewer to low-noise steps for detail refinement.}
    \label{fig:pipeline}
    % \vspace{-0.3cm}
\end{figure}

As illustrated in Figure~\ref{fig:pipeline}, \pare compresses a video Diffusion Transformer (DiT)~\cite{dit} along two axes, \emph{width} (intra-block) and \emph{depth} (inter-block), via spatial-temporal aware width pruning (\S\ref{sec:width}), content-adaptive block routing (\S\ref{sec:routing}), and a progressive training pipeline (\S\ref{sec:training}).

\subsection{Preliminaries}
\label{sec:prelim}

We build on the flow matching framework~\cite{flowmatching, rectifiedflow} for video generation. Given a video latent $x_0 \in \mathbb{R}^{F_l \times H_l \times W_l \times C}$ obtained via 3D VAE compression ($F_l$ temporal slices, spatial resolution $H_l \times W_l$, $C$ channels), the forward process constructs a noisy sample $x_t = (1 - \sigma_t) x_0 + \sigma_t \epsilon$ at time $t \in [0, 1]$, where $\sigma_t$ increases monotonically from $0$ to $1$. A velocity network $v_\Theta$ learns to predict the flow target $(\epsilon - x_0)$ via:
\begin{equation}
    \mathcal{L}_{\text{FM}}(\Theta) = \mathbb{E}_{t, x_0, \epsilon} \left[ \left\| v_\Theta(x_t, t, c) - (\epsilon - x_0) \right\|_2^2 \right],
    \label{eq:fm_loss}
\end{equation}
where $c$ denotes the conditioning signal (text, reference image, or both). Videos are generated by integrating the learned ODE from $t{=}1$ to $t{=}0$.
The velocity network $v_\Theta$ is a DiT with an input projection $g_1$, $N$ sequential transformer blocks, and an output projection $g_2$:
\begin{equation}
    v_\Theta(x_t, t, c) = g_2 \circ B^{N-1} \circ \cdots \circ B^0 \circ g_1(x_t, t, c).
    \label{eq:dit_forward}
\end{equation}
Each block $B^i$ consists of multi-head self-attention (SA, $H_{\text{sa}}$ heads) over the full spatiotemporal token sequence with 3D rotary embeddings~\cite{rope}, multi-head cross-attention (CA, $H_{\text{ca}}$ heads) to text and image conditions, and a feed-forward network (FFN, hidden dimension $d_{\text{ffn}}$). Timestep conditioning enters via adaptive modulation~\cite{dit}. 
% For Wan2.1-14B, $N{=}40$, $H_{\text{sa}}{=}H_{\text{ca}}{=}40$, $d_{\text{ffn}}{=}13824$, and $d{=}5120$, yielding over 16B parameters whose per-step cost motivates the compression strategy described below.

\subsection{Spatial-Temporal Aware Width Pruning}
\label{sec:width}

Attention heads in video DiTs play distinct roles: some process spatial information within temporal slices, while others capture temporal dynamics across slices. Standard magnitude-based pruning overlooks these roles and risks disproportionately removing temporal heads, whose activations tend to be smaller despite their importance for motion coherence. We address this issue through spatial-temporal aware importance scoring and temporal head protection.

\paragraph{Attention Head Pruning.}
Each SA head $h_j^i$ in block $B^i$ contributes $\Delta y_j = W_{O,j}^i \cdot \tilde{\mathbf{v}}_j^i$ to the block output, where $\tilde{\mathbf{v}}_j^i$ denotes the attention-weighted value and $W_{O,j}^i$ is the corresponding slice of the output projection. 
Removing head $j$ induces an output perturbation bounded by $\|W_{O,j}^i\|_F \cdot \|\tilde{\mathbf{v}}_j^i\|_F$. Motivated by the activation-weighted magnitude criterion in WANDA~\cite{wanda}, we use this bound as an importance score that requires neither gradients nor leave-one-out evaluation.
We aggregate importance scores over a calibration set $\mathcal{D}$, using $w(t) = \log(1 + t/T)$ to assign greater weight to high-noise timesteps:
\begin{equation}
    s_{\text{sa}}^{i,j} = \frac{1}{|\mathcal{D}|} \sum_{(x, c) \in \mathcal{D}} \sum_{t=1}^{T} w(t) \cdot \left\| \tilde{\mathbf{v}}_{j}^{i}(x_t, c) \right\|_F \cdot \left\| W_{O,j}^{i} \right\|_F.
    \label{eq:sa_importance}
\end{equation}
Cross-attention importance $s_{\text{ca}}^{i,j}$ is computed analogously; for I2V, text and image contributions are summed ($s_{\text{ca}}^{i,j} = s_{\text{ca,text}}^{i,j} + s_{\text{ca,img}}^{i,j}$). A limitation of magnitude-based scoring is that temporal heads, whose attention spreads across all $F_l$ slices, have lower per-position magnitudes and are systematically underscored.

\begin{figure}
    \centering
    \includegraphics[width=1.0\linewidth]{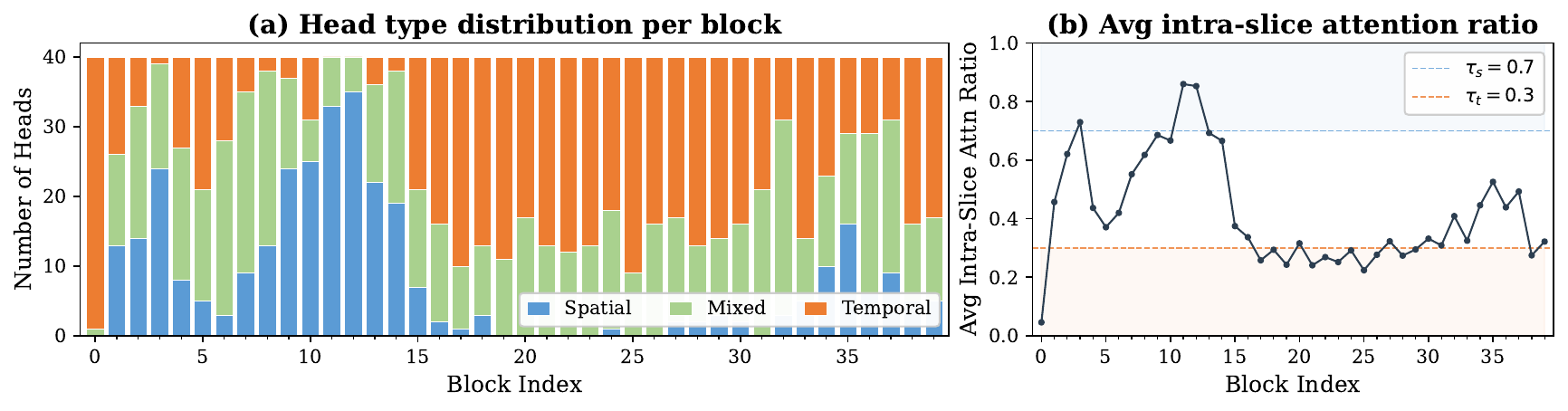}
    \caption{\textbf{Spatial-temporal head analysis of Wan2.1-14B.}
    \textbf{(a)}~Distribution of head types across blocks, classified by the intra-slice attention ratio (Eq.~\eqref{eq:intra_ratio}) with thresholds $\tau_s{=}0.7$ and $\tau_t{=}0.3$.
    \textbf{(b)}~Average intra-slice attention ratio within each block.
    Shallow blocks (3--12) are dominated by spatial heads, whereas mid-to-deep blocks (15--31) are dominated by temporal heads. This variation motivates spatial-temporal aware importance scoring and temporal head protection.}
    \label{fig:head_distribution}
    % \vspace{-0.3cm}
\end{figure}

After VAE compression and patch embedding, a video is represented by $F_l L$ tokens arranged into $F_l$ temporal slices, each containing $L$ spatial tokens. We classify each SA head $h_j^i$ using its \emph{intra-slice attention ratio}, which measures the fraction of attention assigned to tokens within the query's temporal slice:
\begin{equation}
    r_{\text{intra}}^{i,j} = \frac{1}{|\mathcal{Q}|} \sum_{q \in \mathcal{Q}} \frac{\sum_{k \in \mathcal{S}(q)} A_{q,k}^{i,j}}{\sum_{k} A_{q,k}^{i,j}}, \quad \text{type}(h_j^i) = \begin{cases} \text{spatial}, & r > 0.7, \\ \text{temporal}, & r < 0.3, \\ \text{mixed}, & \text{else}, \end{cases}
    \label{eq:intra_ratio}
\end{equation}
where $A_{q,k}^{i,j}$ denotes the post-softmax attention weight from query token $q$ to key token $k$ in head $j$ of block $i$, $\mathcal{S}(q)$ contains the tokens in the same temporal slice as $q$, and $\mathcal{Q}$ is a subset of query positions sampled uniformly across temporal slices. 
As shown in Figure~\ref{fig:head_distribution}, shallow blocks (3--12) are dominated by spatial heads, whereas mid-to-deep blocks (15--31) are dominated by temporal heads.

To correct the magnitude bias, we scale temporal head scores by $\alpha_{\text{temp}}{>}1$:
\begin{equation}
    \tilde{s}_{\text{sa}}^{i,j} = \begin{cases} \alpha_{\text{temp}} \cdot s_{\text{sa}}^{i,j}, & \text{if type}(h_j^i) = \text{temporal}, \\ s_{\text{sa}}^{i,j}, & \text{otherwise}. \end{cases}
    \label{eq:temporal_protection}
\end{equation}
Each block retains $K = \max(K_{\min}, \lceil H_{\text{sa}}(1{-}p) \rceil)$ heads ranked by $\tilde{s}^{i,j}$, where $H_{\text{sa}}$ is the original head count, $p$ is the target pruning ratio, and $K_{\min}{=}8$. 
All blocks retain the same number of heads, but their identities vary across blocks. Temporal protection preserves temporal heads in temporal-dominant blocks, while spatial heads remain favored in spatial-dominant blocks.

\paragraph{FFN Dimension Pruning.}
FFN neurons contribute additively to the output, but importance-based selection alone can retain functionally redundant neurons. We therefore combine importance scoring with diversity-aware selection. Attention heads require calibration to account for data-dependent attention patterns. For FFN neurons, we instead exploit the factored up/down projection structure to define a purely structural importance score:
\begin{equation}
    s_{\text{ffn}}^{i,k} = \|W_{\text{up},k,:}^i\|_2 \cdot \|W_{\text{down},:,k}^i\|_2,
    \label{eq:ffn_importance}
\end{equation}
where $W_{\text{up},k,:}^i$ and $W_{\text{down},:,k}^i$ denote the $k$-th row of the up-projection and the $k$-th column of the down-projection, respectively. To avoid retaining neurons with nearly identical behavior, we define a signature $\mathbf{z}_k^i = [ W_{\text{up},k,:}^i \;;\; W_{\text{down},:,k}^i ] \in \mathbb{R}^{2d}$. We then consider candidates in descending order of $s_{\text{ffn}}^{i,k}$ and skip any candidate whose signature has a cosine similarity greater than $\tau_{\text{ffn}}$ to that of an already selected neuron:
\begin{equation}
    \text{skip neuron } k \;\;\text{if}\;\; \max_{j \in \mathcal{K}} \frac{\mathbf{z}_k^i \cdot \mathbf{z}_j^i}{\|\mathbf{z}_k^i\| \|\mathbf{z}_j^i\|} > \tau_{\text{ffn}},
    \label{eq:ffn_diversity}
\end{equation}
where $\mathcal{K}$ is the set of neurons selected so far. If the number of selected neurons falls short of the target budget after one pass, $\tau_{\text{ffn}}$ is progressively relaxed and the skipped candidates are revisited until the budget is met. This yields a compact FFN where retained neurons are both individually important and functionally diverse. Dimensions are aligned to multiples of 128 for tensor-core efficiency.
After width pruning, we construct the student $v_\Phi$ by extracting the retained weight submatrices from the teacher. The student retains all $N$ blocks, with reduced attention and FFN widths.

\begin{figure}
    \centering
    \includegraphics[width=1.0\linewidth]{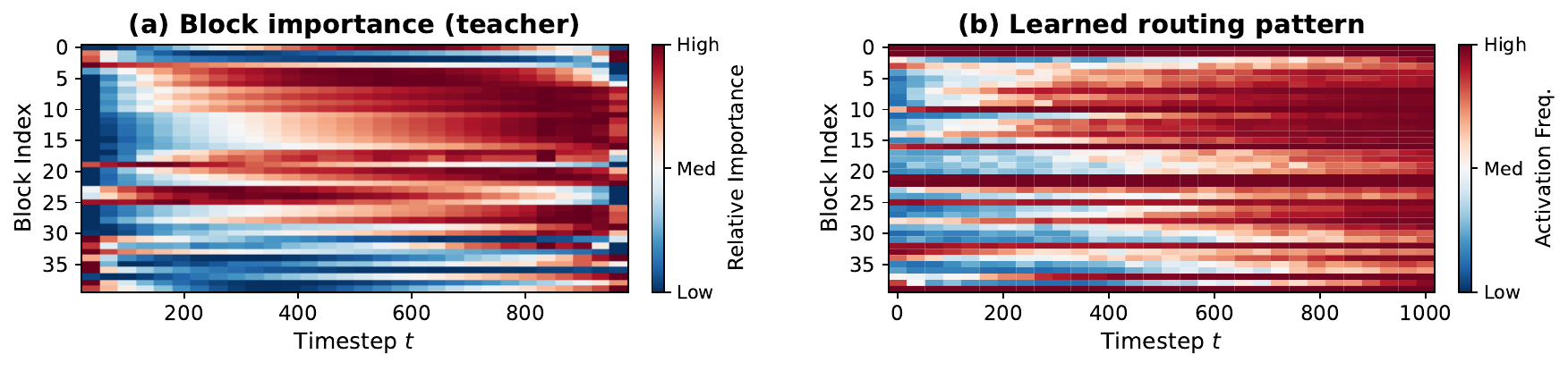}
    \caption{\textbf{From block importance to learned routing.}
    \textbf{(a)}~Teacher block contributions across denoising timesteps, measured by the residual norm $\|B^i(x^{(i)}) - x^{(i)}\|_2$ and independently normalized to $[0,1]$ within each block for visualization. 
    Larger residual norms indicate greater changes to the hidden state. 
    Early and boundary blocks contribute consistently, while middle blocks exhibit timestep-dependent variation.
    \textbf{(b)}~Block activation frequencies under the learned router. 
    Routing reflects the teacher's importance patterns by selectively skipping less important blocks. The budget allocates more blocks to high-noise steps for structure recovery and fewer to low-noise steps for detail refinement.}
    \label{fig:routing_viz}
    % \vspace{-0.3cm}
\end{figure}

\subsection{Content-Adaptive Block Routing}
\label{sec:routing}

Width pruning applies the same compressed architecture to every input and denoising step, leaving variations in denoising difficulty and visual content unexploited. High-noise steps establish global structure, whereas low-noise steps refine details; complex multi-object motion also relies on different blocks than a near-static landscape. Static block pruning~\cite{fastlightgen} fixes the block set and does not account for either variation. We separate block selection into two decisions: a timestep-dependent \emph{budget schedule} determines \emph{how many} blocks to activate, and a timestep- and content-conditioned \emph{router} determines \emph{which} blocks to execute.

To determine how many blocks to activate, we use a linear budget schedule:
\begin{equation}
    K(t) = K_{\min} + (K_{\max} - K_{\min}) \cdot \frac{t}{T},
    \label{eq:budget}
\end{equation}
where $K_{\min}$ and $K_{\max}$ specify the minimum and maximum budgets. The schedule allocates more blocks to high-noise steps ($t \to T$) for recovering global structure, including scene layout, object placement, and motion trajectories, and fewer blocks to low-noise steps for local refinement. Budget values are provided in \S\ref{sec:setup}.

Given this budget, the router $\mathcal{R}_\phi$ determines which blocks to execute by mapping timestep and content embeddings to $N$ block-wise importance logits:
\begin{equation}
    \mathcal{R}_\phi(t, c) = W_3 \, \sigma\!\left( W_2 \, \sigma\!\left( W_1 \, [\text{SinEmb}(t); \, e_c] + b_1 \right) + b_2 \right) + b_3,
    \label{eq:router}
\end{equation}
where $\text{SinEmb}(t)$ denotes a sinusoidal timestep embedding, $e_c$ is a content embedding, and $\sigma$ is the SiLU activation. For I2V, $e_c = \text{CLIP}_{\text{cls}}(c_{\text{img}})$ summarizes the reference image; for T2V, $e_c$ is derived from noisy latent statistics. Timestep conditioning captures which blocks are important at each noise level, while content conditioning enables \emph{per-sample} adaptation. At the same timestep and budget, a fast-motion video may favor temporal-processing blocks, whereas a near-static scene may favor spatial-processing blocks.
We convert the resulting logits $\mathbf{r} = \mathcal{R}_\phi(t, c) \in \mathbb{R}^N$ into a hard binary mask by selecting the $K = K(t)$ highest-scoring blocks:
\begin{equation}
    m_i^{\text{hard}} = \begin{cases} 1, & \text{if } r_i \in \text{top-}K(\mathbf{r}), \\ 0, & \text{otherwise}. \end{cases}
    \label{eq:hard_mask}
\end{equation}
To train the router through this discrete selection, we use the straight-through estimator (STE)~\cite{ste} with the soft surrogate $m_i^{\text{soft}} = \text{sigmoid}(r_i)$:
\begin{equation}
    m_i = m_i^{\text{hard}} + m_i^{\text{soft}} - \text{sg}(m_i^{\text{soft}}),
    \label{eq:ste}
\end{equation}
where $\text{sg}(\cdot)$ denotes the stop-gradient operator. The forward pass uses the hard binary mask, while the backward pass propagates gradients through the sigmoid surrogate, allowing the router to learn from the distillation loss.
The resulting mask controls block execution through a residual blend:
\begin{equation}
    x^{(i+1)} = m_i \cdot B^i(x^{(i)}) + (1 - m_i) \cdot x^{(i)}.
    \label{eq:blend}
\end{equation}
When $m_i = 0$, the block is skipped and $x^{(i+1)} = x^{(i)}$. At inference, skipped blocks are \emph{not executed}, yielding real wall-clock savings. The first and last blocks ($B^0$, $B^{N-1}$) are always active, as they interface with the input and output projections.

\subsection{Progressive Training Pipeline}
\label{sec:training}

Width pruning and block routing compress orthogonal dimensions, but optimizing them jointly from scratch creates a circular dependency: the router cannot learn meaningful block selection until the student produces reasonable features, yet the student cannot compensate for missing blocks until the router makes informed decisions. 
We break this dependency through progressive training, first recovering the width-pruned student and then jointly optimizing the student and router.

\paragraph{Stage I: Width Distillation.}
The width-pruned student $v_\Phi$ is first trained \emph{without routing} via knowledge distillation from the frozen teacher $v_\Theta$. 
The training objective combines four complementary loss terms:
\begin{equation}
    \mathcal{L}_{\text{width}} = w_f \cdot \mathcal{L}_{\text{feat}} + w_t \cdot \mathcal{L}_{\text{TFM}} + w_d \cdot \mathcal{L}_{\text{DFM}} + w_\tau \cdot \mathcal{L}_{\text{temp}}.
    \label{eq:width_loss}
\end{equation}
The \textbf{feature matching loss} $\mathcal{L}_{\text{feat}}$ aligns teacher and student intermediate representations at randomly sampled blocks, using a linear projection $P^i$ when dimensions differ:
\begin{equation}
    \mathcal{L}_{\text{feat}} = \frac{1}{|\mathcal{S}|} \sum_{i \in \mathcal{S}} \left\| P^i(f_\Phi^i) - \text{sg}(f_\Theta^i) \right\|_2^2,
    \label{eq:feat_loss}
\end{equation}
where $\text{sg}(\cdot)$ is the stop-gradient operator. Outlier token positions beyond $3\sigma$ from the mean are masked to stabilize training. 
The \textbf{teacher velocity loss} $\mathcal{L}_{\text{TFM}} = \|v_\Phi - \text{sg}(v_\Theta)\|_2^2$ aligns student predictions with those of the teacher, while the \textbf{data velocity loss} $\mathcal{L}_{\text{DFM}} = \|v_\Phi - (\epsilon - x_0)\|_2^2$ additionally anchors the student to the ground-truth flow target, preventing the student from merely copying the teacher without learning the underlying data distribution. 
The \textbf{temporal consistency loss}~\cite{lv2025dual,habibian2022delta,shen2026chronovision} $\mathcal{L}_{\text{temp}} = \|\Delta_F(v_\Phi) - \text{sg}(\Delta_F(v_\Theta))\|_2^2$ matches inter-slice differences $\Delta_F(\cdot)$ between student and teacher to explicitly supervise motion coherence.
% Loss weights and training details are provided in \S\ref{sec:setup}.

\paragraph{Stage II: Joint Width-Routing Training.}
Starting from the Stage~I checkpoint, we introduce the router $\mathcal{R}_\phi$ and jointly optimize it with the student $v_\Phi$. 
We retain the objective in Eq.~\eqref{eq:width_loss}, but restrict feature matching to \emph{active} blocks ($m_i > 0.5$), excluding skipped blocks from intermediate supervision:
\begin{equation}
    \mathcal{L}_{\text{joint}} = w_f \cdot \mathcal{L}_{\text{feat}}^{\text{active}} + w_t \cdot \mathcal{L}_{\text{TFM}} + w_d \cdot \mathcal{L}_{\text{DFM}} + w_\tau \cdot \mathcal{L}_{\text{temp}}.
    \label{eq:joint_loss}
\end{equation}
% The student and router use separate learning rates---the router, being much smaller ($\sim$470K vs. $\sim$11B parameters), requires a higher rate to learn routing patterns quickly. Both use cosine scheduling with linear warmup.

\paragraph{Stage III: Step Distillation.}
The per-step compressed model from Stage~II is compatible with step distillation methods~\cite{dmd, dmd2}.
Reducing sampling from approximately 50 steps to 4 provides an additional multiplicative speedup on top of the per-step compression achieved in Stages~I and~II.

\paragraph{Compression Analysis.}
We estimate the total speedup by combining width compression, block routing, and sampling-step reduction. After width pruning, block $B^i$ retains a fraction $\rho^i$ of its original FLOPs, and routing activates only $K(t)$ blocks at timestep $t$. Let $C_B$ denote the average FLOPs per block in the unpruned model, $m_i(t) \in \{0, 1\}$ the routing mask defined in Eq.~\eqref{eq:hard_mask}, and $C_{\text{embed}}$ the fixed cost of the input and output projections. The per-step cost is $\mathcal{C}(t) = C_{\text{embed}} + \sum_i m_i(t) \cdot \rho^i \cdot C_B$, giving the following expected speedup:
\begin{equation}
    \text{Speedup} \approx \frac{N}{\bar{\rho} \cdot \bar{K}} \cdot \frac{S_{\text{orig}}}{S_{\text{distill}}} \cdot G,
    \label{eq:speedup}
\end{equation}
where $\bar{\rho}$ denotes the mean width retention, $\bar{K}$ the mean number of active blocks, and $S_{\text{orig}}/S_{\text{distill}}$ the sampling-step reduction factor. The CFG multiplier $G$ equals $2$ when guidance is removed and $1$ otherwise. With $\bar{\rho} {\approx}\, 0.7$ and $\bar{K} {\approx}\, 27$, width pruning and routing yield approximately $2\times$ per-step speedup. Combining this reduction with four-step sampling ($S_{\text{orig}}/S_{\text{distill}} = 50/4$) and $G=2$ gives a projected total speedup of approximately $50\times$.

\section{Experiments}
\label{sec:experiments}

\begin{table}[t]
\centering
\caption{\textbf{Text-to-video generation results on VBench.} We report VBench dimension scores (\%). \textbf{Bold}: best among compressed methods. \underline{Underline}: second best. \colorbox{tablehl}{Highlighted}: our full method.}
\label{tab:main_t2v}
\small
\setlength{\tabcolsep}{2.7pt}
\begin{tabular}{l c c c c c c c c c}
\toprule
\textbf{Method} & \textbf{\#Act.Params} & \textbf{Steps} & \textbf{SubCon} & \textbf{BgCon} & \textbf{Aesth.} & \textbf{Img.Q} & \textbf{Mot.Sm} & \textbf{Dyn.}   & \textbf{Avg.} \\
\midrule
\multicolumn{10}{l}{\textit{Un-accelerated Text-to-video methods}} \\
VideoCrafter2~\cite{chen2024videocrafter2}           &1.4B  &50$\times$2   &\underline{98.16}   &\textbf{98.48}   &66.34   &\underline{68.46}   &98.42   &17.48   &74.56 \\
LTX-Video~\cite{hacohen2024ltx}               &13B  &40   &\textbf{98.42}   &\underline{97.76}   &\underline{66.80}   &\textbf{69.97}   &\textbf{99.07}   &23.93   &75.99 \\
CogVideoX1.5~\cite{yang2024cogvideox}            &5B  &50$\times$2   &92.66   &94.95   &56.24   &62.81   &96.96   &\textbf{46.11}   &74.95 \\
Wan2.1-T2V~\cite{wan2025wan}    &1.3B  &50$\times$2   &96.91   &97.16   &66.26   &67.49   &98.52   &33.89   &\underline{76.70} \\
Wan2.1-T2V~\cite{wan2025wan} (Teacher)    &14B  &50$\times$2   &96.99   &97.49   &\textbf{68.76}   &68.20   &\underline{98.59}   &\underline{36.20}   &\textbf{77.70} \\
\midrule

\multicolumn{10}{l}{\textit{Compression methods on Wan2.1-T2V-14B}} \\
NeoDragon~\cite{karnewar2025neodragon}       &10.5B$_{(25\%\downarrow)}$  &50$\times$2   &\underline{94.33}   &\underline{96.28}   &\textbf{59.50}   &\textbf{69.00}   &\underline{97.44}   &\textbf{31.54}   &\textbf{74.68} \\
ICMD~\cite{wu2024individual}       &9.80B$_{(30\%\downarrow)}$  &50$\times$2   &71.42   &87.46   &29.13   &37.66   &97.20   &\underline{15.35}   &56.37 \\
F3-Pruning~\cite{su2024f3}       &10.5B$_{(25\%\downarrow)}$  &50$\times$2   &\textbf{97.84}   &\textbf{96.50}   &\underline{57.92}   &\underline{57.83}   &\textbf{98.55}   &10.37   &\underline{69.84} \\
\midrule

\multicolumn{10}{l}{\textit{Distillation methods on Wan2.1-T2V-14B}} \\
DMD2~\cite{lightx2v}        &14B  &4   &95.38   &96.16   &68.00   &\textbf{70.54}   &98.36   &\textbf{40.66}   &\textbf{78.18} \\
rCM~\cite{zheng2025large}        &14B  &4   &96.92   &\underline{97.20}   &\textbf{69.87}   &\underline{69.30}   &\underline{98.66}   &32.54   &77.42 \\
AccVideo~\cite{zhang2025accvideo}    &14B  &5   &\underline{97.13}   &\textbf{97.85}   &\underline{69.32}   &62.29   &\textbf{98.85}   &30.06   &75.92 \\
CausVid~\cite{yin2025slow}    &14B  &4   &\textbf{97.34}   &97.17   &68.54   &69.14   &98.36   &\underline{34.17}   &\underline{77.45} \\
\midrule

\rowcolor{tablehl}
\pare (Width-only)      &9.8B$_{(30\%\downarrow)}$  &50$\times$2   &97.36   &\underline{97.50}   &\underline{67.81}   &\underline{71.39}   &\textbf{98.79}   &31.90   &\textbf{77.46}
\\
\rowcolor{tablehl}
\pare (Routing-only)    &9.5B$_{(32\%\downarrow)}$  &50$\times$2   &96.55   &97.21   &66.30   &70.02   &97.76   &\textbf{35.24}   &\underline{77.18}  \\
\rowcolor{tablehl}
\pare &6.7B$_{(52\%\downarrow)}$  &50$\times$2   &\underline{97.56}   &97.37   &66.39   &\textbf{71.55}   &\underline{98.76}   &30.98   &77.10 \\
\rowcolor{tablehl}
\pare + Step Distill. &7.7B$_{(45\%\downarrow)}$  &4   &\textbf{98.75}   &\textbf{98.33}   &\textbf{67.92}   &69.86   &98.74   &28.40   &77.00 \\
\bottomrule
\end{tabular}
\vspace{-0.5cm}
\end{table}

\subsection{Experimental Setup}
\label{sec:setup}

\paragraph{Implementation Details.}
We evaluate \pare on the I2V-14B and T2V-14B variants of Wan2.1~\cite{wan2025wan}, each with 40 DiT blocks, $d{=}5120$, $H_{\text{sa}}{=}H_{\text{ca}}{=}40$, and $d_{\text{ffn}}{=}13824$.
For both tasks, we prune SA, CA, and FFN widths by $30\%$, targeting approximately $30\%$ overall parameter reduction.
Importance scores are computed using 200 calibration samples across 10 timestep bins, with temporal head protection factor $\alpha_{\text{temp}}{=}1.5$.
Head selection within each block follows the spatial-temporal aware scoring in \S\ref{sec:width}, and retained FFN dimensions are aligned to multiples of 128 for tensor-core efficiency.
The router uses budget bounds $K_{\min}{=}20$ and $K_{\max}{=}35$.

We train on 30K video-text pairs with precomputed VAE latents and T5 and CLIP embeddings.
Stage~I (width distillation) and Stage~II (joint width-routing training) each run for 7,000 steps on 4$\times$H200 GPUs using FSDP, and a batch size of 4.
The student and router learning rates are $3{\times}10^{-5}$ and $5{\times}10^{-4}$, respectively, with cosine decay and a 100-step warmup.
Loss weights are $(w_f, w_t, w_d, w_\tau) = (10, 6, 1, 4)$ for I2V and $(10, 6, 1, 8)$ for T2V.
For Stage~III (step distillation), we use the Self-Forcing framework~\cite{huang2025self} and train for 1,200 steps with a learning rate of $2{\times}10^{-6}$.

\paragraph{Baselines.}
We compare against two categories of acceleration methods: 
(1)~\emph{structural compression}, including NeoDragon~\cite{karnewar2025neodragon}, ICMD~\cite{wu2024individual}, and F3-Pruning~\cite{su2024f3}; and
(2)~\emph{step distillation}, including DMD2~\cite{dmd2}, AccVideo~\cite{zhang2025accvideo}, CausVid~\cite{yin2025slow}, and rCM~\cite{zheng2025large}, which reduce sampling steps while retaining the full model.
We also report unaccelerated video generation models for reference.
To isolate the contributions of width pruning and block routing, we include two variants: \emph{Width-only}, the pruned student without routing, and \emph{Routing-only}, the full-width model with block routing.
The unpruned Wan2.1-14B~\cite{wan2025wan} with 50-step sampling serves as the teacher upper bound.

\paragraph{Evaluation.}
We evaluate generation quality using VBench~\cite{vbench}, reporting six dimensions: Aesthetic Quality, Imaging Quality, Motion Smoothness, Dynamic Degree, Background Consistency, and Subject Consistency, together with their normalized average.
Inference uses the FlowUniPC scheduler~\cite{unipc} with 50 sampling steps and a classifier-free guidance scale of 5.0.

\subsection{Quantitative Results}

\begin{table}[t]
\centering
\caption{\textbf{Image-to-video generation results on VBench.} Same evaluation protocol as Table~\ref{tab:main_t2v}.}
\label{tab:main_i2v}
\small
\setlength{\tabcolsep}{2.7pt}
\begin{tabular}{l c c c c c c c c c}
\toprule
\textbf{Method} & \textbf{\#Act.Params} & \textbf{Steps} & \textbf{SubCon} & \textbf{BgCon} & \textbf{Aesth.} & \textbf{Img.Q} & \textbf{Mot.Sm} & \textbf{Dyn.}   & \textbf{Avg.} \\
\midrule
\multicolumn{10}{l}{\textit{Un-accelerated Image-to-video methods}} \\
SEINE~\cite{chen2023seine}           &0.9B  &250$\times$2               &95.02   &95.75   &59.21   &\textbf{70.55}   &96.65   &25.63   &73.80 \\
I2VGen-XL~\cite{zhang2023i2vgen}       &1.4B  &50$\times$2   &\textbf{96.65}   &\textbf{97.27}   &\underline{65.45}   &69.49   &\textbf{98.43}   &20.28   &74.60 \\
DynamiCrafter~\cite{xing2024dynamicrafter}   &1.4B  &50$\times$2   &\underline{95.39}   &95.37   &\textbf{66.38}   &69.09   &97.46   &31.97   &75.94 \\
CogVideoX1.5~\cite{yang2024cogvideox} & 5B &50$\times$2     &94.53   &95.13   &62.20   &69.44   &\underline{98.35}   &34.49   &75.69 \\

% Wan2.2-TI2V~\cite{wan2025wan}  &5B  &50$\times$2            &   &   &   &   &   &   & \\
Step-Video-TI2V~\cite{huang2025step}         &30B  &50$\times$2        &95.36   &\underline{96.12}   &63.74   &68.38   &97.40   &\underline{37.34}  &\underline{76.39} \\
Wan2.1-I2V~\cite{wan2025wan} (Teacher) &16.2B  &50$\times$2   &95.23   &96.05   &65.03   &\underline{70.47}   &98.19   &\textbf{42.54}   &\textbf{77.92} \\
\midrule

\multicolumn{10}{l}{\textit{Compression methods on Wan2.1-I2V-14B}} \\
NeoDragon~\cite{karnewar2025neodragon}    &12.2B$_{(25\%\downarrow)}$  &50$\times$2   &\textbf{93.45}   &\textbf{94.67}   &\textbf{59.12}   &\textbf{67.40}   &\underline{97.02}   &\underline{30.23}   &\textbf{73.65} \\
ICMD~\cite{wu2024individual}       &11.3B$_{(30\%\downarrow)}$  &50$\times$2   &\underline{72.00}   &\underline{86.65}   &30.72   &45.45   &\textbf{98.93}   &12.98   &57.79 \\
F3-Pruning~\cite{su2024f3}       &12.2B$_{(25\%\downarrow)}$  &50$\times$2   &71.93   &85.00   &\underline{35.70}   &\underline{63.05}   &93.97   &\textbf{55.67}   &\underline{67.55} \\
\midrule

\multicolumn{10}{l}{\textit{Distillation methods on Wan2.1-I2V-14B}} \\
DMD2~\cite{lightx2v}        &16.2B  &4   &94.00   &95.00   &\underline{63.74}   &\underline{69.64}   &\underline{97.47}   &\textbf{38.59}   &\underline{76.40} \\
AccVideo~\cite{zhang2025accvideo}    &16.2B  &5   &\textbf{95.62}   &\textbf{96.02}   &63.09   &69.28   &\textbf{97.88}   &18.21   &73.35 \\
CausVid~\cite{yin2025slow}    &16.2B  &4   &\underline{94.32}   &\underline{95.03}   &\textbf{64.87}   &\textbf{70.85}   &97.35   &\underline{38.12}   &\textbf{76.76} \\
\midrule

\rowcolor{tablehl}
\pare (Width-only)      &11.3B$_{(30\%\downarrow)}$  &50$\times$2   &94.86   &95.25   &\underline{63.13}   &\textbf{70.90}   &\underline{98.39}   &\textbf{40.00}   &\textbf{77.09} \\
\rowcolor{tablehl}
\pare (Routing-only)    &10.9B$_{(32\%\downarrow)}$  &50$\times$2   &\textbf{96.50}   &94.38   &\textbf{65.07}   &68.17   &98.33   &\underline{36.86}   &\underline{76.55} \\
\rowcolor{tablehl}
\pare &7.7B$_{(52\%\downarrow)}$  &50$\times$2   &95.34   &\underline{95.36}   &62.12   &\underline{70.57}   &\textbf{98.56}   &35.49   &76.24 \\
\rowcolor{tablehl}
\pare + Step Distill. &8.9B$_{(45\%\downarrow)}$  &4   &\underline{95.60}   &\textbf{95.98}   &62.97   &68.86   &98.32   &33.84   &75.93 \\
\bottomrule
\end{tabular}
\vspace{-0.5cm}
\end{table}

\paragraph{Text-to-Video.}
Table~\ref{tab:main_t2v} reports results on Wan2.1-T2V-14B~\cite{wan2025wan}.
\pare achieves quality close to the teacher while substantially outperforming the structural compression baselines at similar or lower compression ratios.
These include NeoDragon~\cite{karnewar2025neodragon}, which prunes along the depth dimension, and ICMD~\cite{wu2024individual} and F3-Pruning~\cite{su2024f3}, which lack spatial-temporal awareness.
Combining \pare with step distillation maintains generation quality with only four sampling steps, confirming that per-step compression and step reduction provide complementary acceleration.

% \vspace{-0.3cm}

\paragraph{Image-to-Video.}
Table~\ref{tab:main_i2v} shows a similar trend on I2V.
\pare closely preserves teacher quality while halving per-step computation.
It substantially outperforms NeoDragon~\cite{karnewar2025neodragon} despite a greater reduction in active parameters, suggesting that depth-only pruning leaves substantial intra-block redundancy unexploited.
Compared with full-model step distillation methods, \pare achieves quality competitive with DMD2~\cite{dmd2} and CausVid~\cite{yin2025slow} at a fraction of their per-step computational cost.
These results highlight the parameter efficiency of jointly compressing width and depth, rather than relying solely on sampling-step reduction.

\subsection{Qualitative Results}

\begin{figure}[t]
    \centering
    \includegraphics[width=\linewidth]{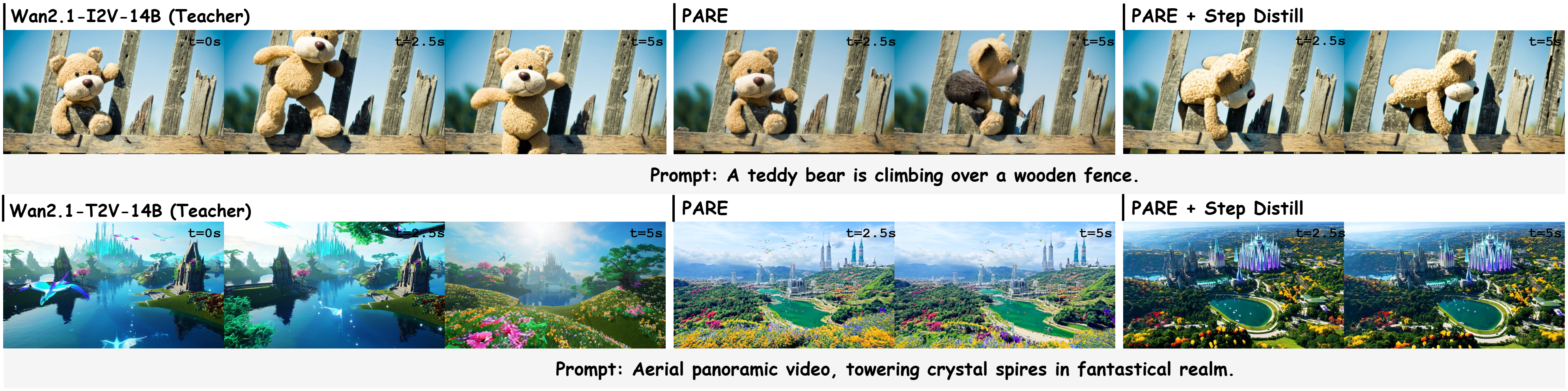}
    \caption{\textbf{Qualitative comparison} between the teacher, \pare, and \pare + Step Distill. }
    \label{fig:qualitative}
    \vspace{-0.3cm}
\end{figure}

Figure~\ref{fig:qualitative} compares representative I2V and T2V results.
For I2V, the teddy bear sequence demands fine texture and coherent motion.
\pare preserves the fur detail and climbing motion, although minor artifacts appear in later frames.
Step distillation resolves these artifacts, suggesting that it provides additional refinement beyond reducing the sampling steps.
For T2V, the aerial fantasy landscape requires coherent global structure.
\pare preserves the composition and color fidelity of the teacher output, with quality maintained after distillation to four sampling steps.

\subsection{Ablation Studies}
\label{sec:ablations}

We ablate the main design choices on I2V following \S\ref{sec:setup}, with additional T2V results for router conditioning.
Highlighted rows denote the proposed configurations.
\vspace{-0.3cm}

% \medskip
\begin{table}[htbp]
\centering
\begin{minipage}[t]{0.49\linewidth}
\vspace{0pt}
\centering
\small
\setlength{\tabcolsep}{4.5pt}
\captionsetup{font=small,skip=4pt,hypcap=false}
\caption{Width pruning at a $30\%$ pruning ratio.}
\label{tab:ablation_width}
\begin{tabular}{l|cccc}
\toprule
Strategy &Img.Q$\uparrow$ &Aesth.$\uparrow$ &BgCon$\uparrow$   & Avg.$\uparrow$ \\
\midrule
Uniform &60.04 &57.22 &92.30  & 71.78 \\
\rowcolor{tablehl} \pare &70.90 &63.13 &95.25  & 77.09 \\
\bottomrule
\end{tabular}

\end{minipage}\hfill
\begin{minipage}[t]{0.49\linewidth}
\vspace{0pt}
\centering
\small
\setlength{\tabcolsep}{4.5pt}
\captionsetup{font=small,skip=4pt,hypcap=false}
\caption{Routing versus random block pruning.}
\label{tab:ablation_routing}
\begin{tabular}{l|cccc}
\toprule
Strategy &Img.Q$\uparrow$ &Aesth.$\uparrow$ &BgCon$\uparrow$   & Avg.$\uparrow$ \\
\midrule
Random &66.98 &58.43 &93.63  &73.08  \\
\rowcolor{tablehl} \pare &68.17 &65.07 &94.38  &76.55 \\
\bottomrule
\end{tabular}
\end{minipage}
\end{table}
\noindent
\textbf{Width pruning strategy.}
Table~\ref{tab:ablation_width} shows that spatial-temporal aware selection preserves quality better than uniform width pruning.
A fixed pruning ratio does not account for the distinct functions of attention heads and FFN neurons.
Temporal head protection counteracts the magnitude bias toward spatial heads, reducing the risk of discarding cross-frame interactions.
Meanwhile, importance-plus-diversity selection avoids allocating FFN capacity to redundant neurons.
Together, these mechanisms help retain complementary spatial and temporal processing within the compressed model.
% \par\medskip

\noindent
\textbf{Learned block selection.}
Learned routing outperforms random block pruning in Table~\ref{tab:ablation_routing} because blocks are not interchangeable across the denoising trajectory.
Random removal can discard transformations that are important at a particular noise level, while retaining less useful ones.
The router learns block scores through the distillation objective, favoring selections that preserve the teacher's predictions.
This interpretation is consistent with Figure~\ref{fig:routing_viz}, where block contributions vary across timesteps, and explains why selecting which blocks to execute matters beyond reducing depth alone.
% \par\medskip
\vspace{-0.3cm}

\begin{table}[htbp]
\centering
\begin{minipage}[t]{0.49\linewidth}
\vspace{0pt}
\centering
\small
\setlength{\tabcolsep}{2pt}
\captionsetup{font=small,skip=4pt,hypcap=false}
\caption{Router conditioning for I2V and T2V tasks.}
\label{tab:ablation_conditioning}
\begin{tabular}{l|cccc}
\toprule
Router input &Img.Q$\uparrow$ &Aesth.$\uparrow$ &BgCon$\uparrow$   & Avg.$\uparrow$ \\
\midrule
I2V-$\mathcal{R}_\phi(t)$ & 69.71 &61.33 &95.42 & 75.40 \\
\rowcolor{tablehl} I2V-$\mathcal{R}_\phi(t, c_{\text{img}})$ & 70.57 &62.12 &95.36  & 76.24 \\
T2V-$\mathcal{R}_\phi(t)$ & 71.27 &65.80 &97.91 &76.84 \\
\rowcolor{tablehl} T2V-$\mathcal{R}_\phi(t, c_{\text{lat}})$ & 71.55 &66.39 &97.37 & 77.10 \\
\bottomrule
\end{tabular}

\end{minipage}\hfill
\begin{minipage}[t]{0.49\linewidth}
\vspace{0pt}
\centering
\small
\setlength{\tabcolsep}{2pt}
\captionsetup{font=small,skip=4pt,hypcap=false}
\caption{Stage~II loss ablations on I2V, with both teacher and data velocity losses always kept active.}
\label{tab:ablation_losses}
\begin{tabular}{cccc|ccc}
\toprule
$\mathcal{L}_{\text{feat}}$ & $\mathcal{L}_{\text{TFM}}$ & $\mathcal{L}_{\text{DFM}}$ & $\mathcal{L}_{\text{temp}}$ &Img.Q$\uparrow$ &Mot.Sm$\uparrow$  & Avg.$\uparrow$ \\
\midrule
$\usym{2717}$ & $\usym{2713}$ & $\usym{2713}$ & $\usym{2717}$ &65.61 &96.47 &72.10 \\
$\usym{2713}$ & $\usym{2713}$ & $\usym{2713}$ & $\usym{2717}$ &70.65 &98.19 &75.34 \\
\rowcolor{tablehl} $\usym{2713}$ & $\usym{2713}$ & $\usym{2713}$ & $\usym{2713}$ &70.57 &98.56 &76.24 \\
\bottomrule
\end{tabular}

\end{minipage}
\end{table}
\noindent
\textbf{Content conditioning.}
Table~\ref{tab:ablation_conditioning} shows that visual conditioning benefits both tasks, with a larger gain on I2V.
Timestep information describes the noise level but cannot distinguish inputs with different visual content; it therefore assigns the same block scores to every sample at a given step.
Content conditioning allows block selection to reflect sample-specific processing needs.
The larger I2V benefit plausibly comes from the reference image's stable appearance and layout cues, whereas noisy latent statistics provide a less direct content signal for T2V.
% \par\medskip

\noindent
\textbf{Distillation objectives.}
Table~\ref{tab:ablation_losses} shows complementary benefits from feature matching and temporal consistency on top of the two velocity losses.
Velocity supervision constrains the final prediction but leaves intermediate representations less directly guided.
Feature matching supplies this guidance at active blocks as the student and router adapt jointly, helping the compressed network retain the teacher's internal representations.
The temporal term adds a different constraint: matching inter-slice differences explicitly couples predictions across frames.
This helps explain its additional benefit beyond matching features and output velocities.
% \par\medskip

% \textbf{Training initialization and budget.}
% Width-distilled initialization outperforms starting from the raw pruned model, consistent with recovering useful student features before learning which blocks to skip.
% Increasing the block budgets improves quality by retaining more transformations along the denoising trajectory, at the cost of additional computation.
% We retain $(K_{\min},K_{\max})=(20,35)$ as the default to favor efficiency.

\subsection{Efficiency Analysis}
\label{sec:efficiency}

We measure batch-one inference in BF16 on one H200 for 81-frame videos at $480\times832$ resolution.

\begin{wraptable}{r}{0.53\textwidth}
\vspace{-\intextsep}
\centering
\small
\setlength{\tabcolsep}{1.5pt}
\captionsetup{font=small,skip=4pt,hypcap=false}
\caption{Measured denoising time (s) and speedup on one H200, relative to the 50-step teacher for I2V and T2V.}
\label{tab:efficiency}
\begin{tabular}{lcccc}
\toprule
 & \multicolumn{2}{c}{I2V} & \multicolumn{2}{c}{T2V} \\
\cmidrule(lr){2-3}\cmidrule(lr){4-5}
Model & Time$\downarrow$ & Speedup$\uparrow$ & Time$\downarrow$ & Speedup$\uparrow$ \\
\midrule
\multicolumn{5}{l}{\textit{(a) 50 sampling steps}} \\
Teacher & 431.15 & $1.00\times$ & 435.14 & $1.00\times$ \\
Width-only & 295.23 & $\textcolor{red}{1.46\times}$ & 313.10 & $\textcolor{red}{1.39\times}$ \\
Routing-only & 344.33 & $\textcolor{red}{1.25\times}$ & 359.78 & $\textcolor{red}{1.21\times}$ \\
\rowcolor{tablehl} \pare & 242.53 & $\textcolor{red}{1.78\times}$ & 258.53 & $\textcolor{red}{1.68\times}$ \\
\midrule
\multicolumn{5}{l}{\textit{(b) 4 sampling steps}} \\
% Model & Time$\downarrow$ & Speedup$\uparrow$ & Time$\downarrow$ & Speedup$\uparrow$ \\
% \midrule
DMD2 (eager) & 17.20 & $\textcolor{red}{25.07\times}$ & 18.36 & $\textcolor{red}{23.70\times}$ \\
DMD2 (compiled) & 15.78 & $\textcolor{red}{27.32\times}$ & 16.38 & $\textcolor{red}{26.57\times}$ \\
\rowcolor{tablehl} \pare + step distill. & 10.81 & $\textcolor{red}{39.88\times}$ & 13.08 & $\textcolor{red}{33.27\times}$ \\
\bottomrule
\end{tabular}
\end{wraptable}
\noindent
The denoising interval includes routing, top-$K$ selection, dynamic dispatch, CFG, and scheduler updates, but excludes conditioning and decoding.
Table~\ref{tab:efficiency} shows that width pruning and block routing provide complementary wall-clock gains: narrower blocks reduce the cost of each executed transformation, while routing reduces how many transformations are evaluated.
Their combination achieves $1.78\times$/$1.68\times$ speedup on I2V/T2V, including routing overhead.
The gap from the nominal $2\times$ compute estimate reflects the deployed schedule, which averages 32 rather than 27 active blocks after timestep shifting and boundary protection.

Width pruning also reduces resident DiT memory by approximately 30\%; routing leaves candidate blocks loaded and therefore adds computational savings without further shrinking resident weights.
With four-step distillation, eager dynamic \pare achieves $1.59\times$/$1.40\times$ denoising speedup over eager full-width DMD2 on I2V/T2V, and remains faster than its compiled counterpart.
These comparisons isolate the benefit of per-step compression after sampling-step reduction.
End-to-end gains are smaller because conditioning and decoding remain outside the compressed denoiser.
Appendix~\ref{app:efficiency} provides the full memory, compilation, end-to-end, and batching measurements.
\par

% \paragraph{Router Behavior Across Timesteps.}

% % TODO: Add figure showing router heatmap (blocks x timesteps)

% Figure~\ref{fig:router_heatmap} visualizes the routing decisions across denoising timesteps. Several patterns emerge: (1) Early blocks (0--5) and late blocks (35--39) are almost always active, serving as input/output processing stages. (2) Middle blocks (15--25) show the most variation, suggesting they encode ``optional'' refinements. (3) At high-noise timesteps ($t > 800$), nearly all blocks are activated, while at low-noise steps ($t < 200$), only 20--25 blocks are used. This validates our timestep-dependent budget design.

% \begin{figure}[t]
%     \centering
%     % \includegraphics[width=\linewidth]{figures/router_heatmap.pdf}
%     \fbox{\parbox{0.95\linewidth}{\centering \vspace{2cm} \textit{Router activation heatmap: blocks (y-axis) vs. timesteps (x-axis). Brighter = more frequently activated.} \vspace{2cm}}}
%     \caption{\textbf{Router activation patterns across denoising timesteps.} The router learns to activate more blocks at high-noise steps and fewer at low-noise steps, with middle blocks being most frequently skipped.}
%     \label{fig:router_heatmap}
% \end{figure}

\section{Conclusion}
\label{sec:conclusion}

We presented \pare, a framework that compresses video DiTs along both width and depth. The two axes demand fundamentally different strategies. Width pruning benefits from understanding the spatial-temporal roles of attention heads and the factored structure of FFN layers, whereas depth reduction benefits from adapting to both the denoising timestep and the visual content. \pare combines spatial-temporal aware width pruning with a learned content-adaptive block router, trained through a progressive pipeline that decouples the two objectives and avoids circular optimization difficulties. On Wan2.1-14B, \pare halves per-step computation for both image-to-video and text-to-video generation while closely preserving quality, and composes naturally with step distillation for acceleration along all three cost axes of video diffusion models.

\paragraph{Limitations.}
Our pruning ratios and router budgets are currently set as global hyperparameters and validated on Wan2.1-14B. Transferring them to architectures with substantially different block designs (e.g., MMDiT, U-ViT) may require re-calibration. The spatial-temporal head classification relies on attention patterns observed during calibration, which could shift under distribution changes such as significantly longer videos or unseen motion categories. \pare also exhibits lower Dynamic Degree scores despite largely preserving Motion Smoothness, suggesting reduced motion amplitude or frequency under compression. Temporal head protection does not fully eliminate this tendency toward lower-motion outputs. 
Additionally, while the progressive training pipeline is more efficient than training from scratch, it still requires access to a moderate-scale dataset and multi-GPU resources, which may limit adoption in resource-constrained settings.

% \paragraph{Societal Impact.}
% By substantially reducing the computational cost of video generation, \pare can lower the energy consumption and carbon footprint associated with large-scale video synthesis, making high-quality generation more accessible to researchers and practitioners with limited hardware budgets. However, the same efficiency gains could also lower the barrier to generating misleading or harmful video content at scale. We encourage the community to pair efficient generation methods with robust watermarking, provenance tracking, and content authentication mechanisms to mitigate potential misuse.

% \section*{References}
% References follow the acknowledgments in the camera-ready paper. Use unnumbered first-level heading for
% the references. Any choice of citation style is acceptable as long as you are
% consistent. It is permissible to reduce the font size to \verb+small+ (9 point)
% when listing the references.
% Note that the Reference section does not count towards the page limit.
% \medskip

% \clearpage
\section{Acknowledgments and Disclosure of Funding}

This work was supported in part by the Australian Research Council under Projects DP240101848 and FT230100549. 
The work was supported in part by the National Key R\&D Program of China (No. 2025YFE0201300).

\bibliographystyle{plain}
\bibliography{refs}

% Source material: author rebuttals tAlTvfKfrn, PLM69hgoS5, vyaddWTVr6.
% OpenReview forum: https://openreview.net/forum?id=WSWLqEZjZV
% Throughput is recomputed from latency to avoid rounding inconsistencies.
\clearpage
\appendix

\section{Practical Inference Efficiency}
\label{app:efficiency}

This appendix supplements the quality evaluation with measured latency, memory, compilation, and batching results. We distinguish denoising-loop latency from end-to-end generation time and report separate benchmark configurations separately.

\subsection{Measurement protocol and fifty-step inference}
\label{app:timing_protocol}

Unless stated otherwise, the measurements in this section use a single NVIDIA H200, BF16 precision, batch size one, and 81-frame videos at $480\times832$ resolution. The synchronized denoising interval includes router inference, top-$K$ selection, dynamic Python block dispatch, classifier-free guidance (CFG), and scheduler updates. It excludes conditioning and video decoding. The teacher, width-only, routing-only, and joint models are evaluated with 50 denoising steps.

\begin{table}[htbp]
\centering
\small
\caption{Fifty-step denoising latency and throughput on Wan2.1. Speedup is relative to the teacher within each task. Throughput is computed as $3600$ divided by the reported denoising time; it excludes the rest of the generation pipeline.}
\label{tab:supp_latency}
\begin{tabular}{llrrr}
\toprule
Task & Model & Time (s)$\downarrow$ & Speedup$\uparrow$ & Videos/h$\uparrow$ \\
\midrule
I2V & Teacher & 431.15 & $1.00\times$ & 8.35 \\
 & Width-only & 295.23 & $1.46\times$ & 12.19 \\
 & Routing-only & 344.33 & $1.25\times$ & 10.46 \\
\rowcolor{tablehl} & \pare & 242.53 & $1.78\times$ & 14.84 \\
\midrule
T2V & Teacher & 435.14 & $1.00\times$ & 8.27 \\
 & Width-only & 313.10 & $1.39\times$ & 11.50 \\
 & Routing-only & 359.78 & $1.21\times$ & 10.01 \\
\rowcolor{tablehl} & \pare & 258.53 & $1.68\times$ & 13.92 \\
\bottomrule
\end{tabular}
\end{table}

Table~\ref{tab:supp_latency} isolates the benefit of per-step compression from sampling-step reduction. Routing provides a further $1.22\times$ speedup on I2V and $1.21\times$ on T2V over the corresponding width-pruned backbone. The combined gains are smaller than a nominal $2\times$ FLOP estimate because realized execution depends on the sampled timesteps and protected blocks, as discussed in \S\ref{app:budget}. These timings include dynamic dispatch rather than assuming that routing is free.

\subsection{Memory and GPU utilization}
\label{app:memory}

Width pruning reduces the resident DiT allocation by physically shrinking its weight tensors. Routing skips execution but keeps the candidate blocks loaded, so it does not provide an additional reduction in resident weight memory. Table~\ref{tab:supp_memory} separates DiT-resident allocation from the denoising peak; neither quantity should be equated with the whole-process memory reported for the caching experiments in \S\ref{app:caching}.

\begin{table}[htbp]
\centering
\small
\caption{Memory and mean GPU utilization for the fifty-step benchmark. Memory is in GiB.}
\label{tab:supp_memory}
\begin{tabular}{llrrr}
\toprule
Task & Model & DiT resident & Denoising peak & GPU util. (\%) \\
\midrule
I2V & Teacher & 31.49 & 36.70 & 99.98 \\
 & Width-only & 22.11 & 26.14 & 99.92 \\
 & Routing-only & 31.49 & 36.71 & 99.94 \\
\rowcolor{tablehl} & \pare & 22.11 & 26.16 & 99.84 \\
\midrule
T2V & Teacher & 27.13 & 32.50 & 99.93 \\
 & Width-only & 19.21 & 24.37 & 99.88 \\
 & Routing-only & 27.13 & 32.51 & 99.86 \\
\rowcolor{tablehl} & \pare & 19.21 & 24.37 & 99.75 \\
\bottomrule
\end{tabular}
\end{table}

The similar width-only and joint-model memory footprints reflect this distinction: width determines the stored parameter volume, whereas routing reduces the executed work. High GPU utilization is maintained across these configurations, but utilization alone is not a measure of useful throughput or kernel efficiency.

\subsection{Dynamic routing and compilation}
\label{app:compilation}

Input-dependent block selection produces multiple execution paths, making a single static whole-model graph less straightforward to reuse. This does not make the compressed operators inherently incompatible with compilation. We evaluate a compiled fixed-route variant and eager execution with fully dynamic routing in a separate T2V benchmark (Table~\ref{tab:supp_compile}). The fixed route executes 29 width-pruned blocks and serves as a compilation diagnostic, not as a replacement for the content-adaptive model.

\begin{table}[htbp]
\centering
\small
\setlength{\tabcolsep}{3pt}
\caption{Compilation diagnostic on T2V. Forward times refer to a conditional DiT forward. The first compiled call includes lazy compilation; the final column measures the 50-step denoising loop in the indicated execution mode.}
\label{tab:supp_compile}
\begin{tabular}{lrrrr}
\toprule
 & \multicolumn{3}{c}{Conditional forward (s)} & Denoising \\
\cmidrule(lr){2-4}
Transformer path & Eager & Compiled & First call & (s) \\
\midrule
Full-width Wan2.1 & 4.287 & 3.930 & 36.85 & 443.2 \\
\pare, fixed 29-block route & 2.257 & 2.021 & 26.91 & 245.8 \\
\pare, dynamic eager & 2.485 & --- & --- & 248.5 \\
\bottomrule
\end{tabular}
\end{table}

Compilation accelerates the fixed-route conditional forward by $1.12\times$. Relative to compiled full-width inference, fixed-route \pare achieves $1.94\times$ faster conditional forwards and $1.80\times$ faster denoising. Fully dynamic eager \pare remains $1.78\times$ faster in denoising than the compiled full-width baseline. Thus, the measured compression benefit persists even when the baseline benefits from compilation. The timings in this diagnostic are reported independently of Table~\ref{tab:supp_latency}.

\subsection{Composition with four-step distillation}
\label{app:four_step}

For the supplementary timing comparison, the full-width four-step DMD2 implementation was adapted from the Self-Forcing framework~\cite{huang2025self}; \pare combines the compressed model with four-step distillation. We additionally benchmark the full-width checkpoint with \texttt{torch.compile}, separating the first invocation from steady execution. \pare uses its deployed eager, dynamic-routing mode.

\begin{table}[htbp]
\centering
\small
\setlength{\tabcolsep}{4pt}
\caption{Four-step inference timings. End-to-end time includes text/image conditioning, denoising, VAE decoding, T5 loading and offloading, and MP4 writing. The compiled steady measurement uses the second sample; first-call timings include lazy compilation. End-to-end times were not reported for the eager full-width baseline.}
\label{tab:supp_four_step}
\begin{tabular}{llrr}
\toprule
Task & Four-step system & Denoising (s) & End-to-end (s) \\
\midrule
I2V & Full-width DMD2, eager & 17.20 & --- \\
 & Full-width DMD2, compiled first call & 52.41 & 74.64 \\
 & Full-width DMD2, compiled steady & 15.78 & 31.83 \\
\rowcolor{tablehl} & \pare + step distillation, eager & 10.81 & 25.52 \\
\midrule
T2V & Full-width DMD2, eager & 18.36 & --- \\
 & Full-width DMD2, compiled first call & 65.85 & 78.89 \\
 & Full-width DMD2, compiled steady & 16.38 & 29.43 \\
\rowcolor{tablehl} & \pare + step distillation, eager & 13.08 & 25.56 \\
\bottomrule
\end{tabular}
\end{table}

The comparison in Table~\ref{tab:supp_four_step} separates the contribution of per-step compression from that of using four sampling steps. \pare is $1.59\times$/$1.40\times$ faster in denoising than the eager full-width four-step baseline on I2V/T2V, and $1.46\times$/$1.25\times$ faster than its compiled steady counterpart. Relative to the respective 50-step teachers in Table~\ref{tab:supp_latency}, denoising is accelerated by $39.88\times$/$33.27\times$. End-to-end gains are smaller because conditioning, decoding, and output operations remain outside the compressed denoiser.

\subsection{Heterogeneous-batch dispatch}
\label{app:batching}

Different samples in a batch can select different block subsets. In the heterogeneous-batch dispatch benchmark, exact routing takes 4.51~s and 9.06~s at batch sizes two and four, respectively, compared with 6.19~s and 12.41~s when every sample executes all 40 width-pruned blocks. This preserves an approximately $1.37\times$ routing gain at both batch sizes. However, latency grows almost linearly with batch size, so these measurements do not show an additional throughput gain from batching. They support small heterogeneous batches; efficient large-batch serving remains an open implementation issue. These dispatch timings are distinct from full denoising-loop measurements.

\section{Routing Behavior and Compression Budgets}
\label{app:routing}

\subsection{Stable and selectively executed blocks}

The 50-step routing traces contain both consistently active blocks and blocks whose execution varies across the trajectory (Table~\ref{tab:supp_routes}). This pattern differs from simply truncating the network: the router preserves a stable set of transformations while selectively using other blocks.

\begin{table}[htbp]
\centering
\small
\caption{Selected statistics from the default 50-step routing traces. The counts describe these traces rather than universal activation frequencies across all inputs.}
\label{tab:supp_routes}
\begin{tabular}{lcl}
\toprule
Task & Blocks active at all 50 steps & Examples of selectively active blocks \\
\midrule
T2V & 15 & Block 33: 9/50; block 14: 22/50 \\
I2V & 23 & Block 7: 14/50; block 12: 22/50 \\
\bottomrule
\end{tabular}
\end{table}

An additional diagnostic over 20 I2V inputs finds 5--9 distinct masks at intermediate timesteps $t\in[200,700]$, with 5--8 blocks varying across inputs. Masks are more similar at the lowest and highest noise levels, where budget constraints and score margins limit changes in selection. Thus, the observed content adaptation affects a bounded subset of blocks: timestep determines the coarse budget, while visual content adjusts which blocks execute within that budget.

Related adaptation methods also depart from uniform treatment of model components and inputs: StructLoRA~\cite{xiao2026not} filters task-irrelevant update directions and coordinates layers, layer-specific prompt fusion~\cite{xiao2026layer} discovers different fusion operators across depth, and ViaPT~\cite{xiao2025visual} generates input-dependent prompts. These methods adapt parameter updates or prompt representations, whereas PARE adapts the executed block subset within a structurally compressed video generator.

\subsection{Nominal compression versus realized execution}
\label{app:budget}

Width pruning removes capacity within every executed block, while routing removes entire residual transformations. Their errors can accumulate, explaining why the combined model can have slightly lower average quality than either single-axis variant at a milder compression setting. Distributing compression across both dimensions avoids placing the full reduction on a single axis. For the nominal retained fractions used in the main comparison, $0.70\times0.68\approx0.48$, corresponding to approximately $52\%$ reduction in the simplified compute model.

The nominal estimate and the measured deployment schedule must be distinguished~\cite{wang2026solving,yang2018netadapt}. The former assumes approximately 27 active blocks out of 40. In the deployed 50-step benchmark, the shifted UniPC timesteps and protected boundary blocks yield an average of approximately 32 active blocks. Equation~\eqref{eq:speedup} should therefore be interpreted as a budget-dependent compute estimate, while Tables~\ref{tab:supp_latency} and~\ref{tab:supp_four_step} report measured latency. 

\section{Diagnostics on Additional Video Backbones}
\label{app:backbones}

We examine whether the motivating head-role heterogeneity and timestep-dependent block importance also occur in HunyuanVideo and HunyuanVideo-1.5~\cite{hunyuanvideo2025}. These are structural and intervention diagnostics on those backbones.

\subsection{Spatial and temporal head roles}

Applying the same intra-slice attention-ratio thresholds as Eq.~\eqref{eq:intra_ratio} identifies spatial, mixed, and temporal heads on both architectures (Table~\ref{tab:supp_head_roles}). The exact proportions differ from Wan2.1, indicating that the head assignments should be calibrated for each backbone rather than transferred unchanged.

\begin{table}[htbp]
\centering
\small
\caption{Head-role counts on additional backbones under the same attention-ratio thresholds. The analyzed head totals are $60\times24$ and $54\times16$, respectively.}
\label{tab:supp_head_roles}
\begin{tabular}{lrrrr}
\toprule
Backbone & Spatial & Mixed & Temporal & Total \\
\midrule
HunyuanVideo & 63 & 328 & 1,049 & 1,440 \\
HunyuanVideo-1.5 & 17 & 141 & 706 & 864 \\
\bottomrule
\end{tabular}
\end{table}

\subsection{Block importance along real trajectories}

We perform leave-one-block-out interventions on four real 50-step denoising trajectories for each backbone. Table~\ref{tab:supp_block_rank} reports Spearman correlations of the resulting block-importance rankings. Rankings are relatively consistent across prompts at the same denoising stage, but change substantially between early and late stages. The results support timestep-dependent block selection beyond Wan2.1, while leaving the quality and efficiency of a fully trained cross-backbone \pare system to future evaluation.

\begin{table}[htbp]
\centering
\small
\caption{Spearman correlations of block-importance rankings on additional backbones. Cross-prompt ranges compare rankings at the same stage; the last column compares early and late denoising.}
\label{tab:supp_block_rank}
\begin{tabular}{lcc}
\toprule
Backbone & Cross-prompt, same stage & Early versus late \\
\midrule
HunyuanVideo & $0.910$--$0.946$ & $0.209$ \\
HunyuanVideo-1.5 & $0.770$--$0.889$ & $-0.188$ \\
\bottomrule
\end{tabular}
\end{table}

\section{Comparison with Caching and Dynamic Image DiTs}
\label{app:caching}

\subsection{DeepCache deployment comparison}

We benchmark the DeepCache~\cite{deepcache} paths on Wan2.1 and HunyuanVideo-1.5 using one H200 and BF16 precision. This HunyuanVideo inference uses $848\times480$ resolution, 121 frames, and CFG scale 6 for the 50-step full-inference and caching configurations. Settings are matched within each backbone; the \pare + step-distillation rows use four steps. Table~\ref{tab:supp_deepcache} reports observed process memory, which includes allocations beyond the resident DiT weights and differs from the memory scope in Table~\ref{tab:supp_memory}.

\begin{table}[htbp]
\centering
\small
\setlength{\tabcolsep}{3.5pt}
\caption{Deployment measurements for caching and four-step \pare. Memory denotes observed process memory in GiB.}
\label{tab:supp_deepcache}
\begin{tabular}{llrrr}
\toprule
Backbone/task & System & Denoising (s) & End-to-end (s) & Memory \\
\midrule
HunyuanVideo-1.5 & Full inference & 442 & 471.47 & $\approx44.3$ \\
 & DeepCache & 225 & 261.10 & $\approx66.9$ \\
\midrule
Wan2.1 T2V & Full inference & 435.14 & 457.98 & $\approx33.54$ \\
 & DeepCache & 242.41 & 263.15 & $\approx111.39$ \\
\rowcolor{tablehl} & \pare + step distill. & 13.06 & 22.89 & $\approx25.97$ \\
\midrule
Wan2.1 I2V & Full inference & 431.15 & 459.91 & $\approx38.59$ \\
 & DeepCache & 241.55 & 268.91 & $\approx118.19$ \\
\rowcolor{tablehl} & \pare + step distill. & 10.52 & 25.86 & $\approx27.82$ \\
\bottomrule
\end{tabular}
\end{table}

On HunyuanVideo-1.5, DeepCache provides $1.96\times$ denoising and $1.81\times$ end-to-end speedup, while increasing observed process memory from approximately 44.3 to 66.9~GiB. Across three prompts and two seeds, its mean end-to-end speedup is $1.830\times$. On Wan2.1, caching similarly reduces latency but requires substantially more process memory in the tested implementation.

The approaches make different deployment trade-offs. DeepCache avoids retraining by reusing features across adjacent denoising steps, whereas \pare incurs training cost to reduce computation within each step. Four-step sampling leaves fewer opportunities for cross-step reuse; the measurements demonstrate \pare's composition with step distillation, but do not measure DeepCache combined with four-step distillation. Table~\ref{tab:supp_deepcache} is therefore a deployment comparison. 

\subsection{Relationship to DyDiT}

DyDiT~\cite{dydit} and \pare both learn to allocate computation adaptively across denoising steps and inputs, but differ in selection granularity. DyDiT uses timestep-conditioned gates to select attention heads and MLP channel groups, together with content-dependent token routing that bypasses selected MLP updates. \pare instead combines permanently pruned widths with whole-block routing conditioned on timestep and content, adapting execution depth while retaining dense token processing within active blocks. Thus, DyDiT exploits within-layer and spatial redundancy, whereas \pare removes entire attention and FFN transformations when a block is skipped, at the cost of coarser selection. For video generation, \pare further distinguishes spatial and temporal head roles during pruning and supervises inter-slice differences during distillation, explicitly addressing temporal information preservation. The methods therefore share a conditional-computation principle but make different trade-offs between fine-grained adaptivity, compact parameterization, and video-specific temporal modeling.

\section{Motion Preservation}
\label{app:motion_limits}

The Dynamic Degree degradation in Tables~\ref{tab:main_t2v} and~\ref{tab:main_i2v} is an important quality limitation. Teacher-to-\pare scores decrease from 36.20 to 30.98 on T2V and from 42.54 to 35.49 on I2V, while Motion Smoothness remains close to the teacher. This pattern is consistent with reduced motion amplitude or frequency rather than primarily increased flicker. A plausible explanation is that aggressive channel and block compression favors more stable, lower-motion trajectories. Temporal-head protection addresses the under-selection of temporal heads, but does not eliminate this motion under-generation.

% \newpage
% \input{checklist.tex}

\end{document}